\documentclass[12pt]{article}

\usepackage[cp1250,cp1251,latin1]{inputenc}
\usepackage[ukrainian,croatian,german,russian,american]{babel}
\usepackage{graphicx}
\usepackage{tipa}
\usepackage{natbib}

\def\RU{\selectlanguage{russian}\inputencoding{cp1251}\selectfont}
\def\US{\selectlanguage{american}\inputencoding{latin1}\selectfont}
\def\UA{\selectlanguage{ukrainian}\inputencoding{cp1251}\selectfont}

\US

\title{Statistical Parameters of the Novel {\sl Perexresni ste\v{z}ky}
({\sl The Cross-Paths}) by Ivan Franko}

\author{Solomija Buk and Andrij Rovenchak\footnote{Corresponding author's e-mail: andrij@ktf.franko.lviv.ua}\\
Ivan Franko National University of Lviv, Ukraine}
\date{}

\begin{document}

\maketitle

\begin{abstract}
In the paper, a complex statistical characteristics of a Ukrainian novel
is given for the first time. The distribution of word-forms with respect to their size is
studied. The linguistic laws by Zipf--Mandelbrot and Altmann--Menzerath
are analyzed.
\end{abstract}

\section{Introduction}
Year 2006 is the 150th anniversary of
Ivan Franko (1856--1916), the prominent Ukrainian writer, poet, publicist, philosopher,
sociologist, economist, translator-polyglot and the public figure.
His incomplete collected works were published in 50 volumes \citep{Franko:1976}.
With this name the notion of national identity in the Western Ukraine is connected.
Franko's works have intensive plot and interesting topic.
In this paper, we make an analysis of the novel{\UA\it Ïåðåõðåñí³ ñòåæêè}
({\sl The Cross-Paths}, referred also as {\sl The Crossroads}
in Encyclop\ae dia Britannica).

The events of the novel unfold at the turn of the 20th century.
This is a
story of young lawyer Eugenij Rafalovy\v{c} coming to the provincial town in Halytchyna (Galicia)
to continue his practice. He meets there his former teacher, Stalski, telling to Eugenij about
his matrimonian life, telling, in particular, that he does not talk at all to his wife for ten (!)
years as a punishment.
The point is that Stalski appears to be married to Regina, Eugenij's Jugendliebe\ldots\
This novel is about lawlessness and justice, meanness and nobleness,
consciousness and subconsciousness.
Social motives, psychologism, love and tragedy are tangled here in an intricate way.
The novel was translated into French %by Ginette Maksymovytch
\citep{Franko:1989}, see
also some excerpts in \cite{Anthologie:2004}, and into Russian %by U\v{s}akov
\citep{Franko:1956}.

The present paper is the first attempt in Ukrainian linguistics to
make a comprehensive quantitative study of a particular work of art
using modern techniques. Previous word-indices of Ukrainian
writers \citep{vashhenko:1964,zhovtobrjux:1978,Kovalyk:1990,lukjanjuk:2004}
were compiled manually, their aim was to establish the number of
occurrences of a particular word, not to make an analysis of such
data.

Small efforts in the quantitative study of the works written by Ivan Franko
were made recently for his fairy-tales, see \citep{holovatch&palchykov:2005}.

The present study is based on the frequency dictionary compiled by the authors
on the basis of the edition \citep{Franko:1979} using the principles consistent with those
described in our recent paper \citep{buk&rovenchak:2004}. We have also analyzed the main differences
of this edition comparing to the first (separate)
one \citep{Franko:1900}.

\vspace *{-0.2cm}
\section{Basic Principles of the Text Analysis}
We consider a token as a word in any form (a letter or alphanumeric sequence between two spaces),
irrespective to the language. Thus, `1848', {\UA`60-èé'}, `\S136' were treated as one token.

We have partially restored the use of letter{\UA\it ´} \textipa{[\textscriptg]} eliminated from the Ukrainian
alphabet in 1933 during Stalin's rule as a step toward removing the differences between
Ukrainian and Russian orthography. The letter{\UA\it ã} was left to denote
both \textipa{[\texthth]} and \textipa{[\textscriptg]} sounds having, however, a sense-distinguishing
role:{\UA\it ãí³ò} `oppression' versus{\UA\it ´í³ò} `wick',{\UA\it ãðàòè} 'to play'
versus{\UA\it ´ðàòè} `bars', etc.
The letter{\UA\it ´} was rehabilitated in the Ukrainian orthography in 1993,
but in a much narrower extent.

We have tried to restore the use of this letter using the edition \citep{Franko:1900} and
following modern Ukrainian orthographical tendencies.
First, in the proper names,{\UA\it Ðå´³íà} `Regina', {\UA\it Âà´ìàí} `Wagman',
{\UA\it Ðåññåëüáåð´}\  `Resselberg'.
Second, in loan words from Polish, German, Latin:%
{\UA\it ´ðàòóëþâàòè} $\leftarrow$ gratulowa\'c,
{\UA\it ´åøåôò} $\leftarrow$ Gesch\"aft,
{\UA\it ìîð´} $\leftarrow$ Morgen (a measure of area),
{\UA\it àáíå´àö³ÿ} $\leftarrow$ abnegation, etc.
And, of course, in those words which are now traditionally written with{\UA\it ´}:
{\UA\it ´àíîê}, {\UA\it ´àòóíîê}, {\UA\it ´ðàñóâàòè}, {\UA\it ´ðóíò}, etc.

\vspace *{-0.2cm}
\subsection{Euphony}
In Ukrainian, some words appear in different phonetic variants caused by
the `phonetic environment' (i.e., the notion of euphony, cf.\ Polish {\it w/we}, {\it z/ze},
Russian {\RU\it ñ/ñî}, {\RU\it ê/êî}, also English indefinite article
{\it a/an} or initial consonants mutations in Irish).
They are initial {\UA \it â/ó, â³ä/â³äî, ç/³ç/ç³/çî, ³/é} and respective prepositions
and conjunctions), final {\UA\it-ñÿ/ñü}).
Such word variants were joined under one (the most frequent) form.
Instead, the vernacular variants are given separately. For instance,
{\UA\it àäóêà(í)ò} and{\UA\it àäâîêàò} (`advocate'),
{\UA\it ïåðåãðàô} and{\UA\it ïàðàãðàô} (`paragraph'),
the second form in the examples is a normative one.

\subsection{Homonyms}
The problem of homonyms is one the most complicated problems slowing down the process of
automatic text processing. This is connected with a very high frequency of auxiliary
parts of speech having in Ukrainian (as well as other Slavic languages) the same form.
For instance, in the text under investigation one has 1956 occurrences of{\UA\it ùî},
distributed as follows: 1360 -- conjunction (`that', `which'), 495 -- pronoun (`what'),
101 -- particle. The token{\UA\it à} occurs 1065 times as a conjunction (`and', `but'),
33 times as a particle and 6 times as an interjection.
The token{\UA\it ÿê} is found 389 times as a conjunction (`as'), 125 -- as and adverb (`how'),
55 -- as a particle. Note, however, that the translations are very approximate due to
a wide range of the word meanings.
The `full-meaning' words occupy a bit lower ranks:{\UA\it ìàòè} appears 35 times as the verb
`to have' in Infinitive versus 4 occurrences as the noun `mother' in Nominative Singular.

While the above examples are standard and expectable for Ukrainian, we have also met some
specific parallel forms:{\UA\it í\'àéìèòè}, the noun `hireling' in Plural Nominative, and
{\UA\it íàéì\'èòè}, the verb `to hire' in Infinitive; {\UA\it ãóñò\'\i}, the adjective `dense'
in Singular Accusative Feminine,
and {\UA\it ã\'óñò³} the noun (in fact,{\UA\it ´\'óñò³}) `taste' in Singular Genitive.

The analysis of the homonyms could not be fully made in an automatic way,
even the contextual analysis is not sufficient. Therefore, a manual control was necessary.

What is interesting, the problem of homonyms appears even in a small subset of
words written in the Latin script:
we had to distinguish Latin and German {\it in},
German definite articles {\it die} (Plural and Feminine),
Latin {\it maxima} (adjective, Feminine from {\it maximus} and noun, Plural from {\it maximum}).

\section{Statistical Data}
The text size $N$ is 93,885 tokens. In the novel, forty five tokens are alphanumeric,
208 are written in the Latin script (in German (87), Latin (55), Polish (38), French (14),
Czech (9), Yiddish (4) languages, once the letter `S' is referred to describe the form of
a river), all the remaining are Ukrainian ones.

The number of different word-forms  is 19,391.
The number of different words (lemmas) -- vocabulary size $V$ -- is 9,962.

Mean word length is 4.83 letters and mean sentence length is 9.7 words.

Vocabulary richness (the variety index) calculated as the relation of the number of words
to the text size equals $V/N=0.106$.

Vocabulary density is calculated as the ratio of text size and the vocabulary size
$N/V=9.42$. Otherwise said, a new word is encountered as every 9--10 words are read.

The number of hapax legomena $V_1$ is 4,902 making up thus 49.2 per cent of the vocabulary
and 5.22 per cent of the text. These parameters are also known as the exclusiveness indices
of text and vocabulary, respectively.

The concentration indices are connected with the number of words having
the absolute frequency equal and higher than 10: $N_{10}=74,692$ for the text and
$V_{10}=1,127$ for the vocabulary. The concentration indices are therefore
$N_{10}/N=79.6\%$ for the text and $V_{10}/V=11.3\%$ for the vocabulary.

The main feature of the edition \citep{Franko:1900}
influencing the statistical parameters of the investigated novel in comparison to the modern
text \citep{Franko:1979} is the usage of the verbal reflexive particle{\UA\it -ñÿ}.
In modern Ukrainian, it is written together with the respective verb, unlike the orthographical
rules of 1900 (cf. also the shortened variant{\UA\it -ñü} written in one word in both older
and modern texts).
In the novel, this particle is used 2496 times in 1485 different verbal forms.
This frequency would correspond to the second (!) highest rank, after
{\UA\it ³/é} `and' (3204).
Such a result correlates with, e.g., modern Polish, where the corresponding
word {\it si\textpolhook{e}} also belongs to the most frequent ones \citep{PWN}, see Table~1.

\section{Distributions and Linguistic Laws}

\begin{figure}[h]
\centerline{\includegraphics[scale=0.3]{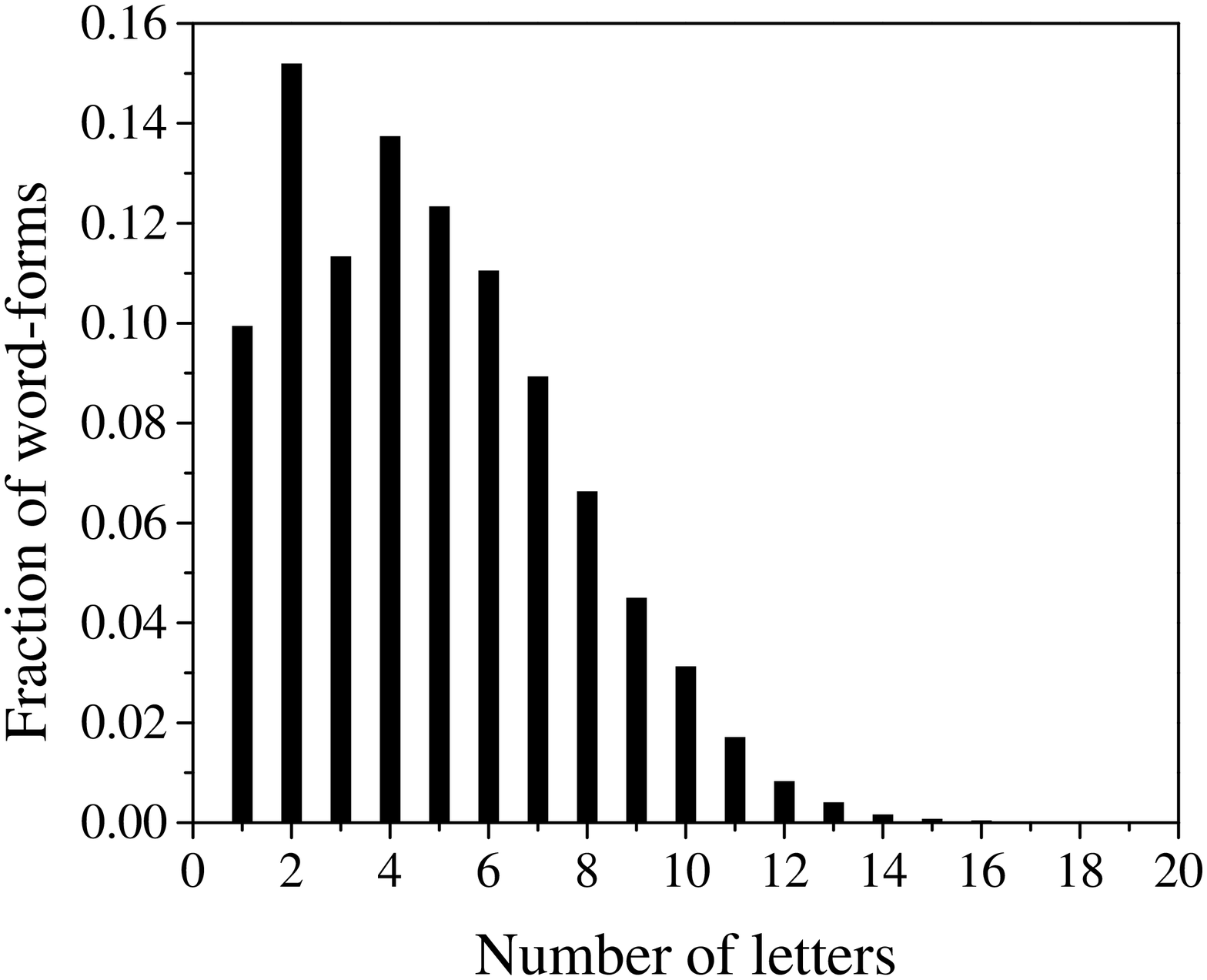}
%\qquad
\hfill
\includegraphics[scale=0.3]{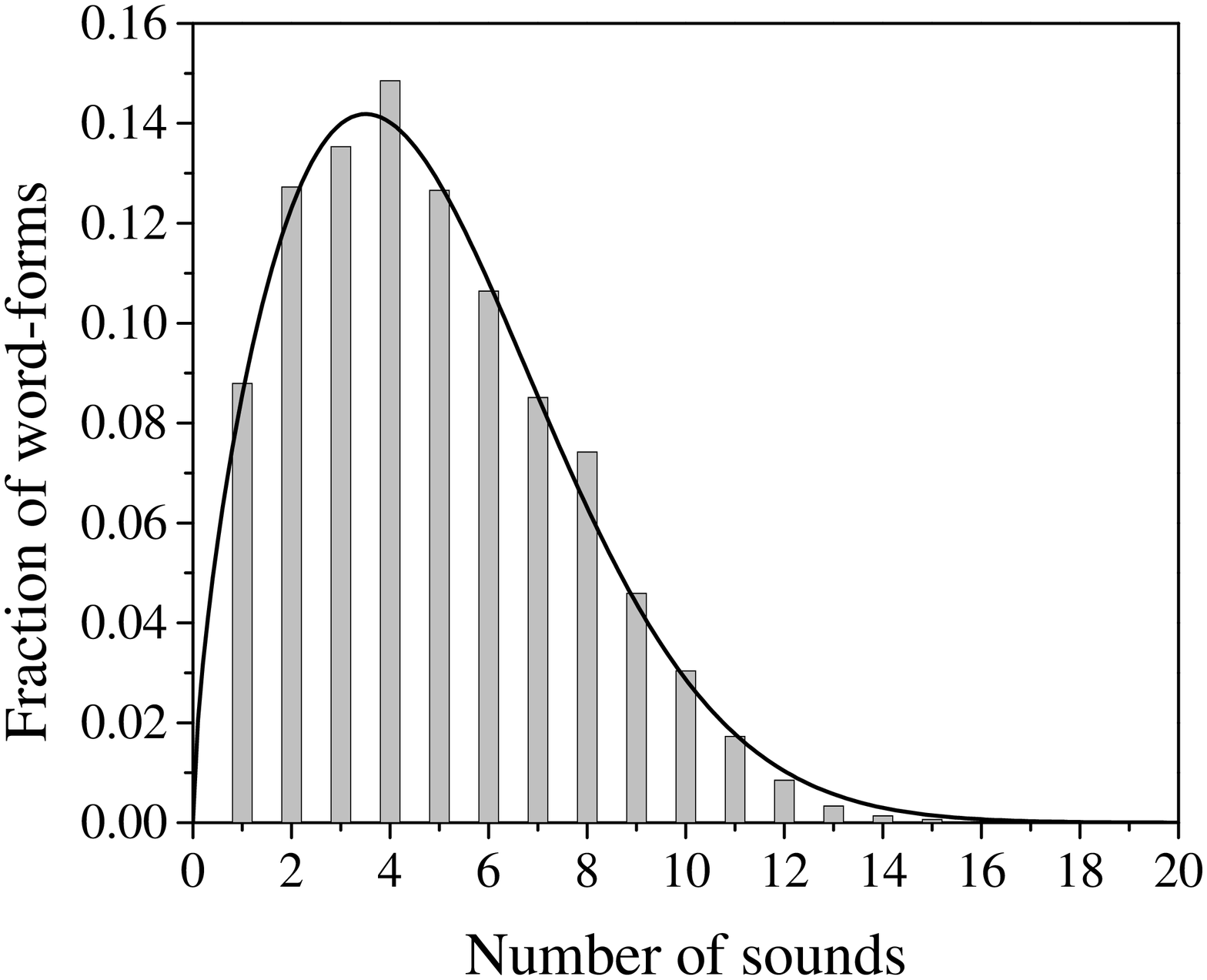}}
\centerline{\hfill(a)
\hfill\hfill
(b)\hfill}
\caption{The distribution of word-forms (fraction of unity, vertical axis) with respect
to the number of constituting letters (a) and sounds (b).}
\label{figLetters}
\end{figure}

We have analyzed the distribution of word-forms with respect to the number of letters and found
that such a dependence has two maxima, see Fig.~\ref{figLetters}(a). As the size of our sample is quite large,
this fact can signify that some other unit must be considered as a proper, or natural one.
A phoneme (sound) and a syllable appeared to be an appropriate alternative.

The dependence between the fraction of word-forms $W$ containing exactly $\varphi$ phonemes
can be approximated by the following (empirical) formula:

\begin{equation}
W=A\varphi^b\,e^{-\alpha \varphi^2}, \qquad
A={2\alpha^{(b+1)/2} \over \Gamma\left(b+1\over2\right)},
\end{equation}
where the value of $A$ is obtained from the normalization condition
\begin{equation}
1=\int_0^\infty A\varphi^b\,e^{-\alpha \varphi^2}\,d\varphi.
\end{equation}

The fitting parameters are as follows: $b = 0.6347,\ \alpha=0.02579$,
see Fig.~\ref{figLetters}(b).
The results regarding fitting in this work were obtained using the nonlinear least-squares
Marquardt--Levenberg algorithm implemented in the GnuPlot utility, version 3.7 for Linux.

If a syllable is used as a length unit, we have utilized the formula similar
to Altmann--Menzerath law \citep{altmann:1980} but with the argument shifted by unity:
\begin{equation}
W=B(s+1)^{d}\,e^{-\gamma (s+1)}, \qquad B={\gamma^{\ d+1}\over \Gamma(d+1)}.
\end{equation}
In the above formula, $W$ is the fraction of word-forms containing
exactly $s$ syllables. The reason to introduce the shift is a high
frequency of non-syllabic words (particles{\UA\it á, æ},
prepositions{\UA\it â, ç}, conjunction{\UA\it é}), which were not
treated as proclitics, in contrast to, e.g., the approach by
\citeauthor*{grzybek&altmann:2002}
(\citeyear{grzybek&altmann:2002}) for similar Russian words. We
have put the length of such words to be zero. Thus, the
distribution function has to be non-zero at the origin ($s=0$).

\begin{figure}[h]
\centerline{\includegraphics[scale=0.3]{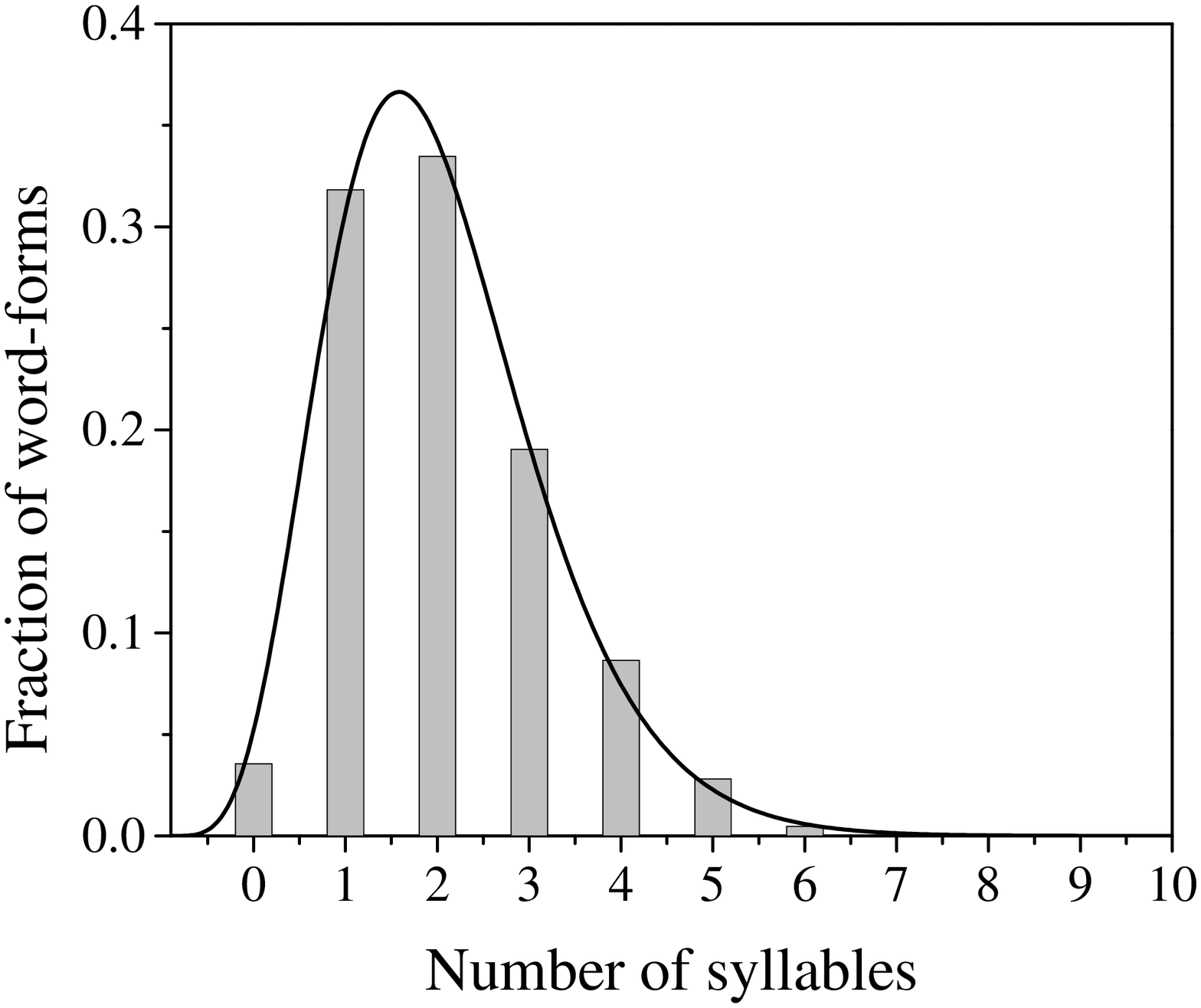}
\qquad\includegraphics[scale=0.3]{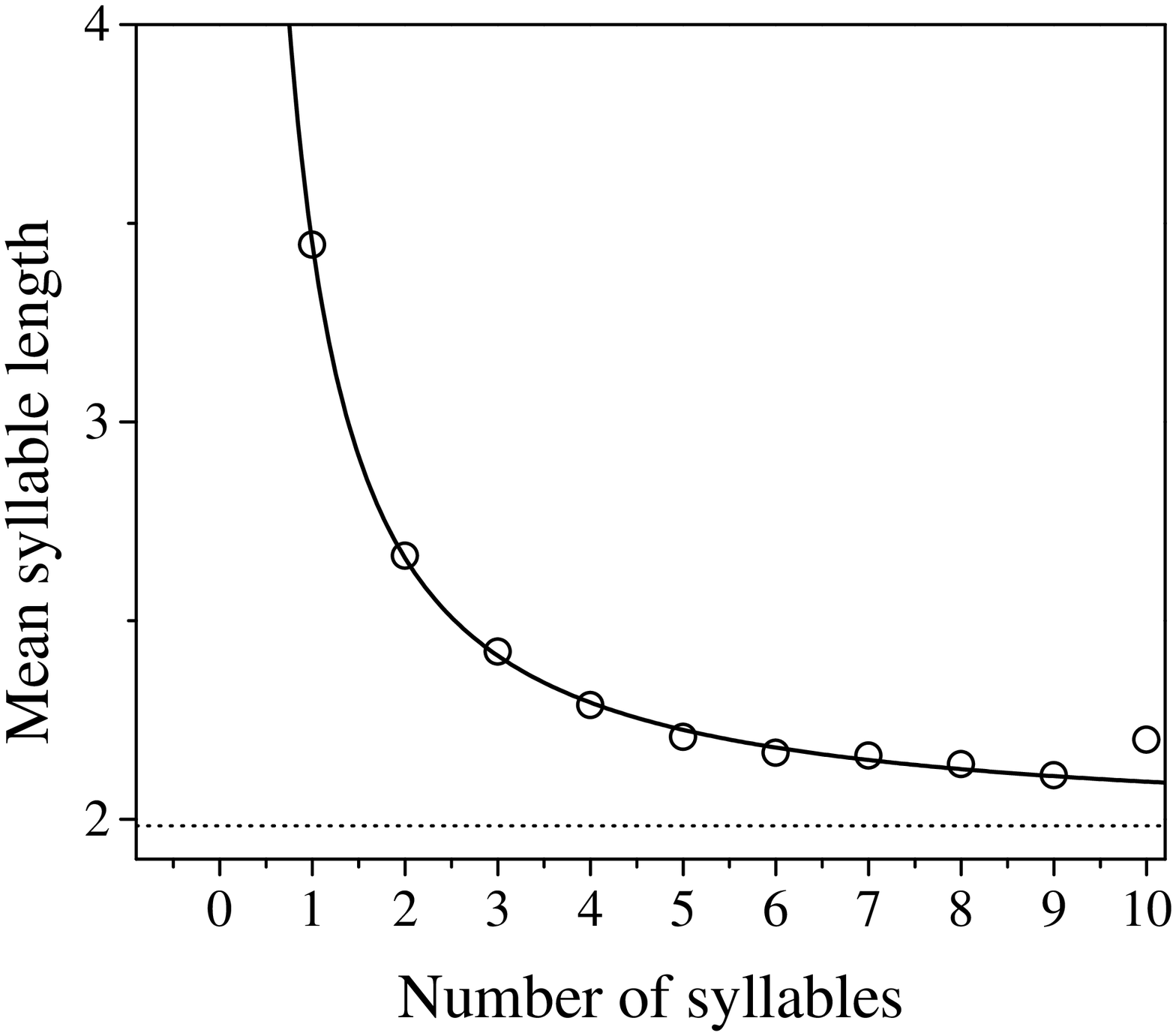}}
\centerline{\hfill(a)
\hfill\hfill
(b)\hfill}
\caption{The distributions regarding syllabic structure of the words:
the fraction of word-forms with respect to constituting syllables (a),
the evidence of Menzerath's law (the right-most point was excluded from the fit
due to poor statistical reliability).}
\label{figSyllables}
\end{figure}

The fitting parameters are as follows: $d = 5.805,\ \ \gamma = 2.245$,
see. Fig.~\ref{figSyllables}(b).

In order to check the validity of Menzerath's law
we have also studied the dependence of the mean syllable length $M$ on the word length $s$
(measured in syllables).
We have used the formula
\begin{equation}
M=M_\infty + B\,s^c.
\end{equation}
The constant $M_\infty$ denotes a possible asymptotic value of the mean syllable length
in a very long word, the exponent $c$ is a negative number. In this way, we also obtain
an infinite value of the syllable length for non-syllabic words ($s=0$).
The fitting parameters are as follows: $M_\infty = 1.984,\ \ B=1.464,\linebreak c = -1.119$.

The form $M=As^b\,e^{cs}$ (see, e.g., \cite{koehler:2002}) appeared to give a poorer fit,
leading in particular to large mean syllable length of long words due to the exponential
increase.

\begin{figure}[h]
\centerline{\includegraphics[scale=0.35]{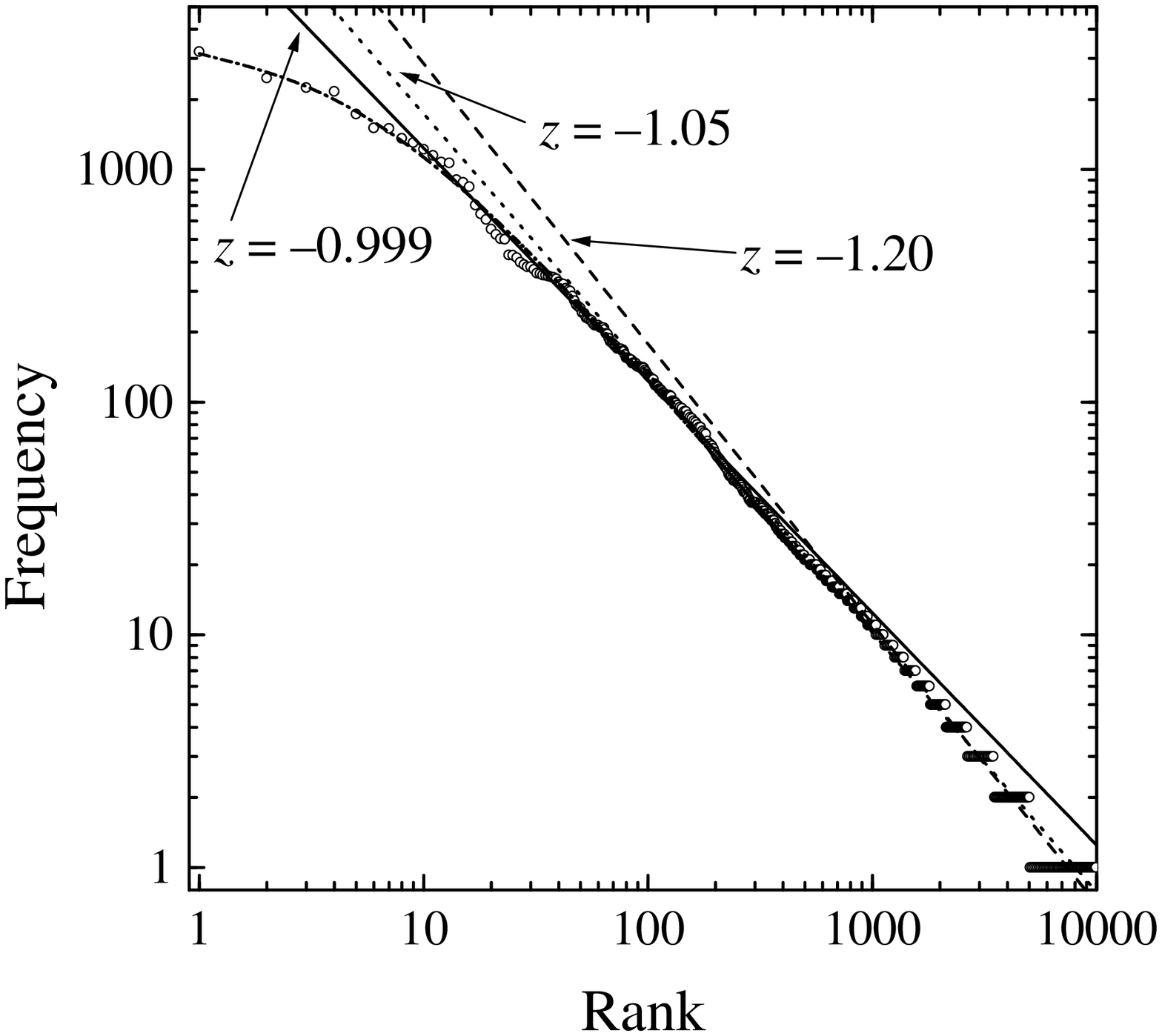}
\qquad\
\includegraphics[scale=0.35]{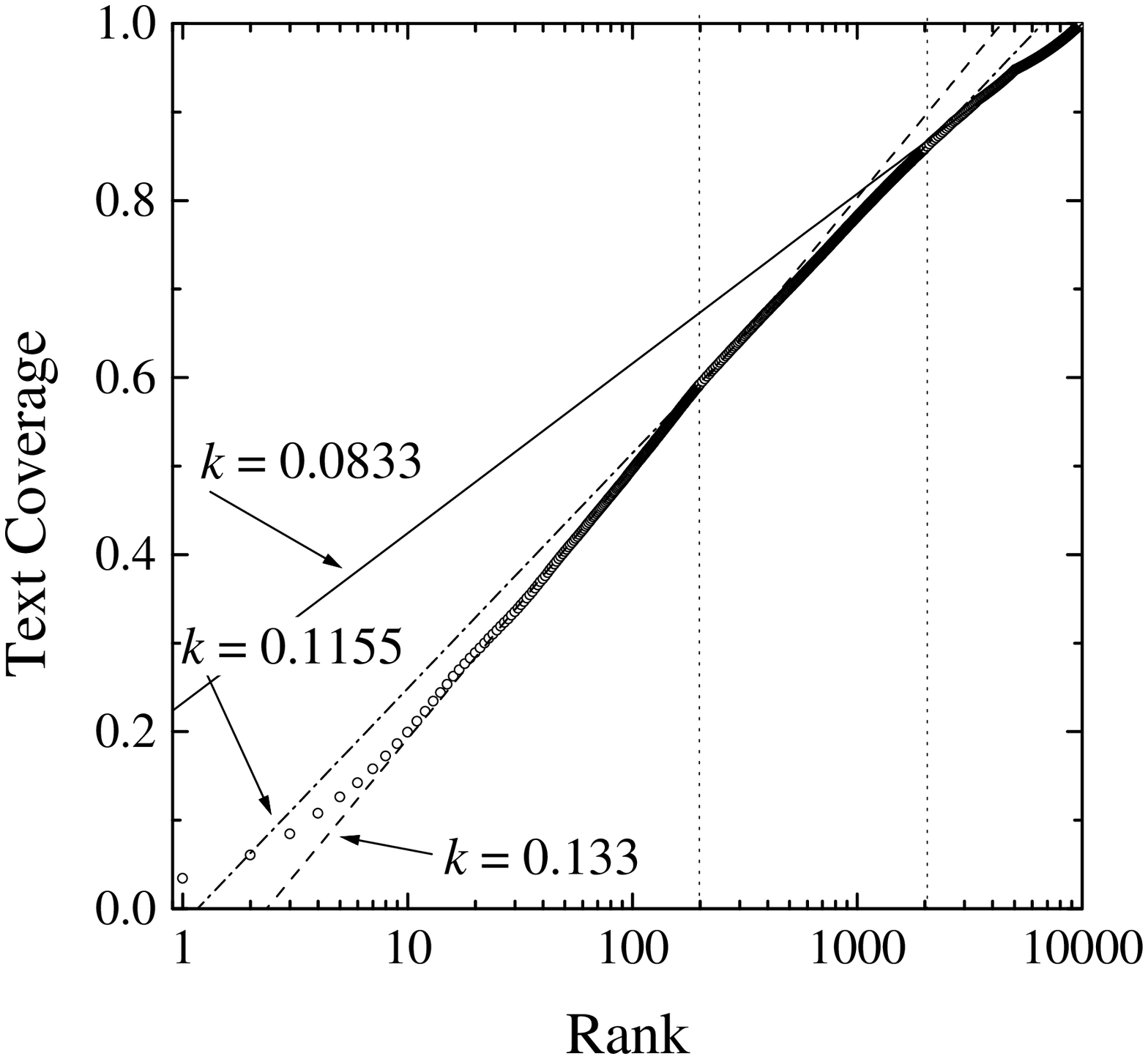}}
\centerline{\hfill(a)
\hfill\hfill
(b)\hfill}
\caption{The transition to different regimes in Zipf's law (a) and
text coverage (b). In (a) the dashed-dotted line is the fit to the Zipf--Mandelbrot law.}
\label{figZipfMandelbrot}
\end{figure}

%\subsection{Zipf--Mandelbrot law}
We have calculated the parameters of Zipf's law fitting our frequency data in different ranges
of ranks.
One has the word frequency $F$ connected with its rank $r$ via simple relation:
%\begin{equation}
$F(r)=A/r^z$.
%\end{equation}
The values of the exponent $z$ can be related to the different types of vocabulary.
Visually in Fig.~\ref{figZipfMandelbrot} we can see three such rank domains:
$10<r<200$ ($z=-0.999$), $200<r<1000$ ($z=-1.05$), $r>1000$ ($z=-1.20$).

The parameters of the Zipf--Mandelbrot law $F(r)=A/(r+C)^b$ were also calculated for the whole
rank domain: $A=25000$, $b=1.14$, $C=5.2$. See
Fig.~\ref{figZipfMandelbrot}(a) for the results.

%\subsection{Text coverage}
The portion of text $T$ covered by first $r$ ranked words can be fitted by the dependence
$T(r)=k\ln r + T_0$. While for $10<r<200$ the growth of the text coverage is characterized by
$k=0.133$, it slows a bit for $200<r<2,000$ with $k=0.1155$ and
even more for the larger values of $r$, $k=0.0833$ for $r> 2,000$.
See Fig.~\ref{figZipfMandelbrot}(b) for details.

\section{Comparison}

To complete our paper, we adduce the comparison of the top-ranked words in five different
languages (see Table~1 in the next page).
The Ukrainian text is the novel under consideration, the English is {\sl Ulysses} by James
Joyce \citep{Ulysses}, the Japanese is {\sl Kokoro} by Natsume
S\=oseki \citep{Ohtsuki}, Russian corresponds to the vocabulary of Lermontov \citep{Lermontov}, and Polish
is from PWN (\citeyear{PWN}).

As expected, the majority of these words are auxiliary parts of
speech, irrespective to language. What is interesting, in the text
of a particular writing (Ukrainian, English and Japanese examples)
some common features are found. Namely, the names of the
characters have very high frequency allowing them to reach the
highest ranks, together with addresses{\UA\it ïàí}, {\it Mr.},
\includegraphics[scale=1.0]{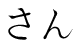}.\linebreak Also,
the nouns denoting human body-parts are quite frequent, in
particular `hand' is not found only in the Polish list (the reason
is probably a comparatively large fraction of journalistic texts
in the PWN corpus). These phenomena require additional
interlingual studies.

%\newpage

\renewcommand{\baselinestretch}{.75}

\noindent
\setlength\tabcolsep{4pt}
\begin{table}
\caption{The top-ranked words, in the right columns the relative frequency is given in per cent.\protect\\}
\includegraphics[scale=1.0]{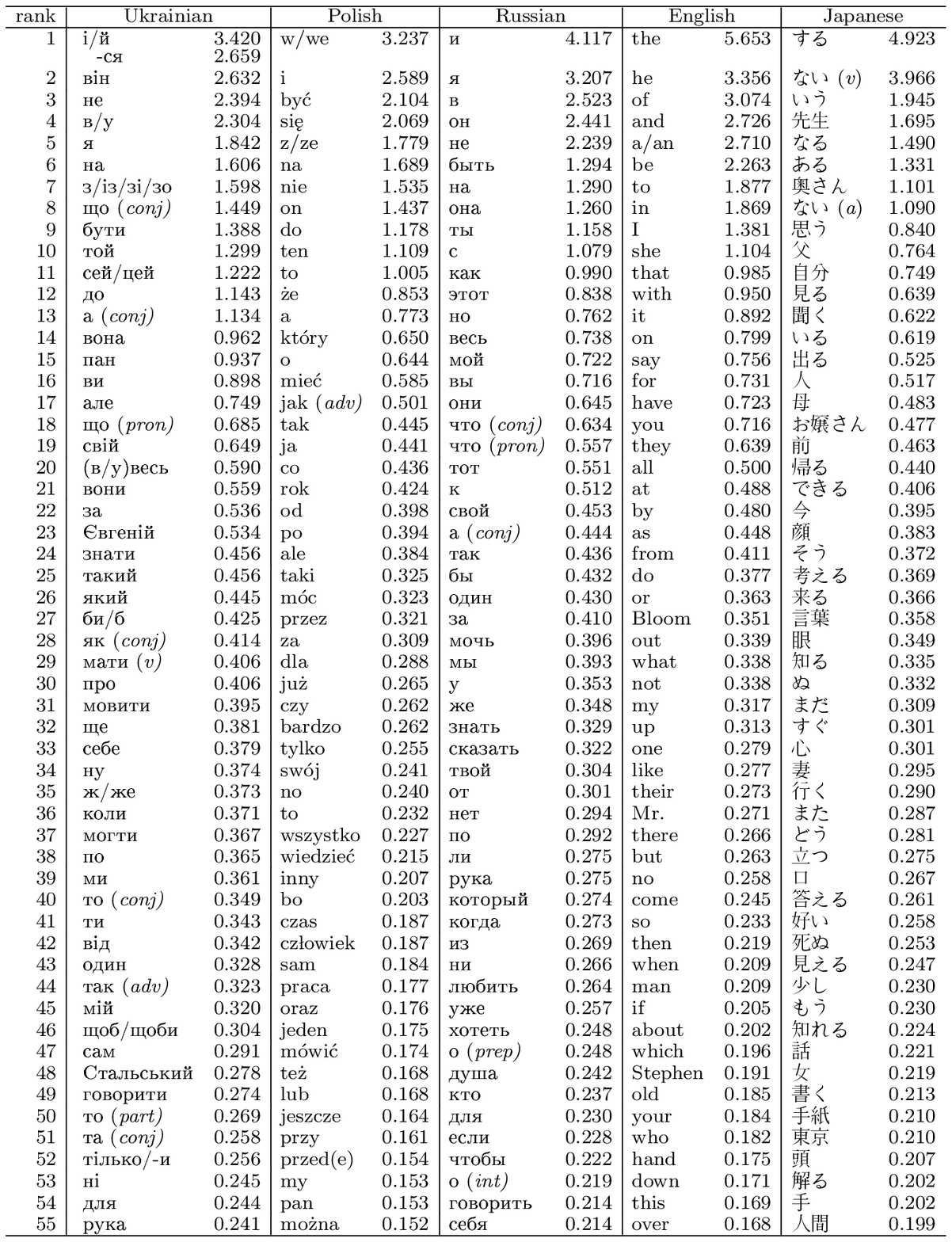}
\end{table}

\renewcommand{\baselinestretch}{1.0}

\newpage

\bibliographystyle{chicago}

\end{document}